\documentclass{article}

\usepackage[numbers]{natbib}            %

\usepackage{arxiv}
\usepackage[utf8]{inputenc} 
\usepackage[T1]{fontenc}    
\usepackage{hyperref}       
\usepackage{url}            
\usepackage{booktabs}       
\usepackage{amsfonts}       
\usepackage{nicefrac}       
\usepackage{microtype}      
\usepackage[table,xcdraw]{xcolor}
\usepackage{color}
\usepackage{graphicx} 
\usepackage{subfigure}

\usepackage{amssymb}
\usepackage{enumerate}
\usepackage[shortlabels]{enumitem}
\usepackage[english]{babel}
\usepackage[utf8]{inputenc}
\usepackage{algorithm}
\usepackage[noend]{algpseudocode}

\title{Slot-based Image Augmentation System for Object Detection}

\author{
  Yingwei Zhou\thanks{GitHub:
  \href{}{https://mercurise.github.io/}} \\
  School of Electronics and Computer Science\\
  University of Southampton\\
  Southampton, SO17 1BJ \\
  \texttt{yz39g15@soton.ac.uk} \\
}

\begin{document}
\maketitle

\begin{abstract}
Object Detection has been the significant topic in computer vision. 
As the continuous development of Deep Learning, many advanced academic and industrial outcomes are established on localising and classifying the target objects, such as instance segmentation, video tracking and robotic vision. 
As the core concept of Deep Learning, Deep Neural Networks (DNNs) and associated training are highly integrated with task-driven modelling, having great effects on accurate detection. 
The main focus of improving detection performance is proposing DNNs with extra layers and novel topological connections to extract the desired features from input data. 
However, training these models can be a computational expensive and laborious progress as the complicated model architecture and enormous parameters. 
Besides, dataset is another reason causing this issue and low detection accuracy, because of insufficient data samples or difficult instances. 
To address these training difficulties, this thesis presents two different approaches to improve the detection performance in the relatively light-weight way. 
As the intrinsic feature of data-driven in deep learning, the first approach is ``slot-based image augmentation" to enrich the dataset with extra foreground and background combinations. 
Instead of the commonly used image flipping method, the proposed system achieved similar mAP improvement with less extra images which decreases training time. 
This proposed augmentation system has extra flexibility adapting to various scenarios and the performance-driven analysis provides an alternative aspect of conducting image augmentation.
\end{abstract}


\section{Introduction}
Object detection is a fundamental research area to answer where are the objects and what they are. 
It consists of two key sub-problems of localisation and classification.
Despite its simple conception, many advanced computer vision tasks are based on precisely acquired object location and categories by object detection methods, such as instance segmentation, video tracking and autonomous driving \citep{Chen2015ICCV, lee2018memory, schulter2017deep}. 
Therefore detection quality has a great impact on these advanced vision tasks. 
The main contributions of this thesis are introducing two novel aspects to improve the detection performance which are light-weight, flexible to extend and less computational cost. 
To be specific, an image augmentation system called ``slot-based image augmentation" is proposed and experiments have shown its improvements for small object detection and huge potential capacity in augmenting images. 
Moreover, it is implemented in an image pre-processing fashion, no need for training and easy to extend for various scenarios. 
This approach highlights the significance of data-source, addressing the very intrinsic of data-driven machine learning approaches. 
Besides, this work has found failure of commonly used image augmentation methods that they might damage the mAP of detection DNNs. 
\newline 

In practice, the limitation of available data is a common case and data augmentation methods such as flipping, rotating and tuning brightness, are applied to get through the bottleneck of lacking training data. 
These methods are conducted as manual image transformation without task-oriented features. 
Although these methods contribute partial increase for image classification accuracy, they are not that beneficial for object detection. 
In some cases, augmentation makes the model performance worse and related results are presented in this Chapter. 
The main reason of the failures is the differences in target learn-able features between image classification and object detection.  
In addition, object detection requires extra location and contextual features rather than the object feature alone as in classification task. 
Moreover the manual image transformation based augmentation are not guaranteed to augment sufficient extra learn-able features especially for small scale objects.   
\newline 

Therefore, the above discussion draws the motivation that an image augmentation approach is desired specially for object detection task to provide extra classification and localisation features. 
Regarding the methodological gap between object detection and regular augmentation methods, the slot-based image augmentation is proposed to generate images with more learn-able object detection related features by adding extra combinations of foreground objects and background images.

\section{Related Works}

\begin{figure}[!htbp]
  \centering
    \includegraphics[width=0.8\linewidth]{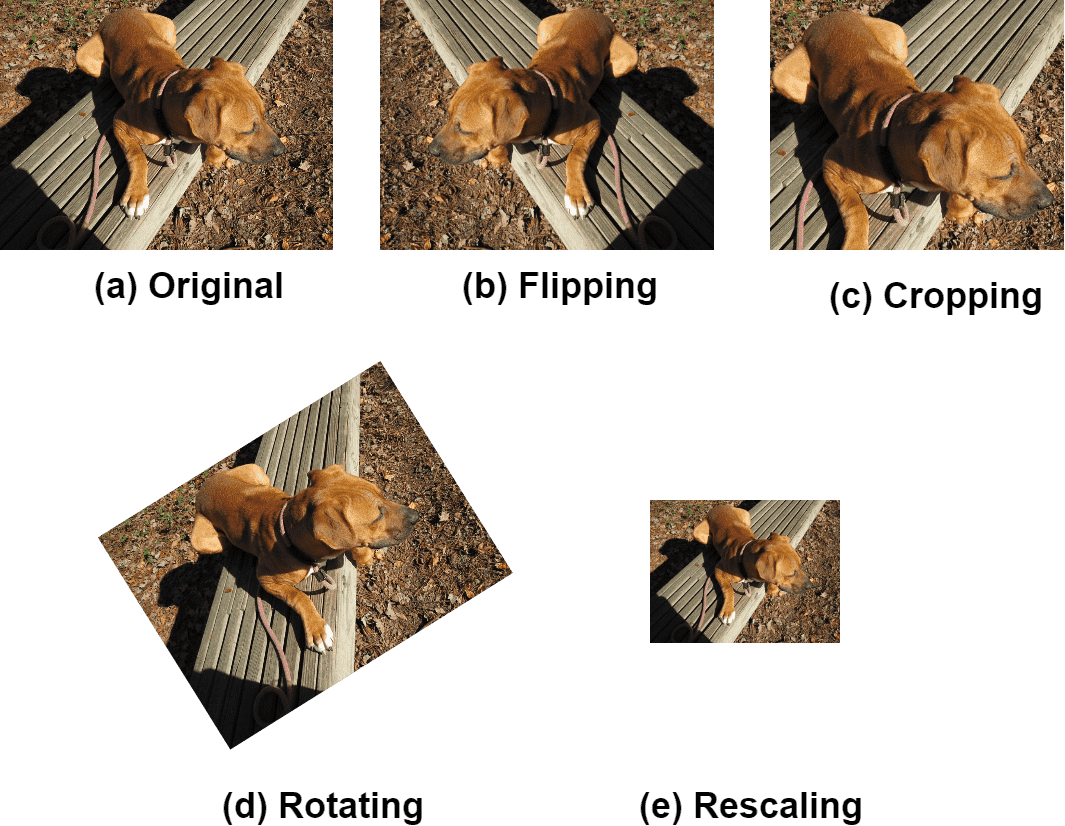}  
  \caption{\textbf{The Examples of Commonly Used Image Augmentation Methods}.}
\label{fig:common-augmentation}
\end{figure}
Commonly used image augmentation approaches include manual image transformations such as flipping, rotating, cropping and re-scaling as shown in Figure \ref{fig:common-augmentation}. 
Because images are represented as matrix in computer vision, these transformations are conducted by linear transformation on matrix and many toolboxes support these transformations such as OpenCV \cite{opencv_library}, Scikit-image \cite{van2014scikit} and recently released ``Population Based Augmentation" by \cite{ho2019population}. 
Besides of these transformations on image size or positions, another alternative approach is the colour-wise augmentation including brightness, contrast and saturation, which are supported by many deep learning frameworks: TorchVision \cite{paszke2017automatic}, mxNet \cite{chen2015mxnet}, etc. 
In summary, most of the augmentation works are based on simple image transformation and there are not many other augmentation approaches featured with performance analysis or adaptive augmentation. 
\newline 

Besides these augmentation methods, there are some works on inserting objects into background image to generate new images. 
\cite{dvornik2018modeling} proposed a machine learning based method to search suitable positions to insert the objects. 
These positions are ranked by the performance evaluation of the applied object detection models. 
Among these works, \cite{singh2018analysis} highlights the importance of scales and \cite{DBLP:journals/corr/abs-1902-07296} conducted similar works on small object detection.
\newline

Instead of these augmentation related works, the main stream focuses on modifying DNN architecture to improve detection performance by mixing CNN feature maps of different layers. 
For the small object detection, the contextual information is highlighted thus various DNN architectures are proposed to utilise these features as possible such as original feature pyramid pooling by \cite{he2015spatial}, the relation based DNN architecture by \cite{hu2018relation} and contextual refinement methods by \cite{Chen_2018_ECCV}.
\newline

These architectural modifications methods achieved desired detection improvements however, added extra computation costs and less flexible for transferring the trained models to different scenarios. 
Inspired by these proposed novel methods and impressive implementation works, the proposed slot-based image augmentation method is lightweight, flexible for transfer learning and highlights a general way of augmenting images for object detection.

\section{slot-Based Image Augmentation System}
\label{section:slot-based-augmentation-system}

\subsection{System Introduction}
\begin{figure}[!htbp]
  \centering
    \includegraphics[width=0.8\linewidth]{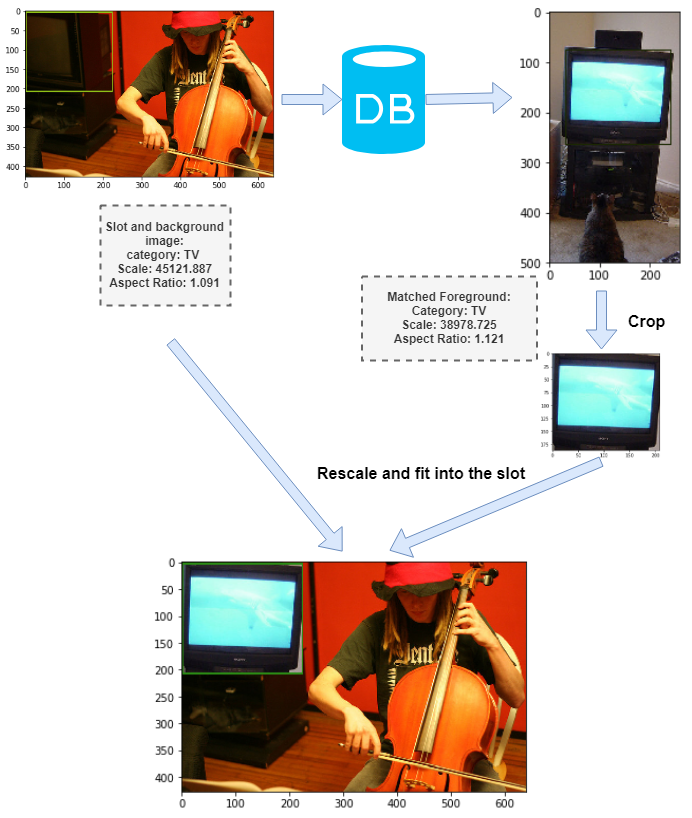}  
  \caption{Generating one image by fitting one TV slot with a valid foreground object which has the similar aspect ratio and scales.}
\label{fig:demo-same}
\end{figure}

The slot-based augmentation system is built on the fundamental element called ``slot" which is a generalised conception from the isolated foreground objects. 
In other words, a slot is the replaceable position which is initially the isolated foreground object in an image and the slot-based image augmentation produces extra images by substituting foregrounds enriching various combinations of foreground and backgrounds. 
Furthermore detecting objects is the process of recognising foreground objects from its background utilisation the features learnt from training images. 
Therefore the slot substitution enriches the combinations especially for the unbalanced dataset in which there are not sufficient objects of a certain category. 
In addition, the slot based augmentation is easy to customise and highly flexible for different scenarios such as substituting objects of the same category, which creates extra learn-able ``foreground and background" correlation features for the target category objects. 
Figure \ref{fig:demo-same} example of augmentation. 
It is also a potential way to change the bounding box style slot substitution to an instance wise substitution as the instances for segmentation tasks are represented as polygons. 
\newline 
 
\section{System Implementation}
 
Regarding the conception of slots, the two highlighted features are foregrounds and isolation. 
On one hand, foregrounds assign slot location as the foregrounds which tightly include objects inside the bounding box and causes limited corruption for other objects or background while substituting them. 
On the other hand, the isolation selects individual objects that have no insertion with any other objects and no damage to other objects when replacing them with other ones. 
Hence a slot is a rectangle area of the background image with the same position as an isolated original object and causes minor effects while substitute it. 
\newline 

Despite the simple definition, it takes several steps to conduct the augmentation and the system is designed in an end-to-end fashion making it flexible to modify and extend with more functionalities, which are presented in the following section. 
\subsection{System Design and Pipeline}
\label{subsection:augmentation-pipeline}
As introduced above, slot is the central component of the augmentation system which is required to be isolated and possessing the foreground object locations. 
Dataset for object detection includes images and annotation files that specify object location as coordinates and object information such as category and scales. 
Thus foreground object locations are capable to be directly extracted from the annotation files while isolation requires extra steps to extract them. 
The progress of selecting isolated objects is simplified as finding individual rectangle bounding boxes that have no insertion with any others in the same image. 
As bounding boxes are represented with coordinates, a heuristic method is applied to solve this problem by comparing a set of coordinates as in Figure and the rules are shown below: 
\newline 
For two bounding boxes represented by the coordinates of the top-left and bottom-right vertexes as $bbox1$: $[x_{11},\ y_{11}, \ x_{12},\ y_{12}]$ and $bbox2$: $[x_{21},\ y_{21}, \ x_{22},\ y_{22}]$, where $(x_{11},\ y_{11})$ and $x_{21},\ y_{21}$ are top-left vertices while the bottom-right ones are $x_{12},\ y_{12}$ and $x_{22},\ y_{22}$. So $bbox1$ and $bbox2$ are overlapped if and only if satisfying all the following conditions and the isolated slots are decided by the complementary cases: 
\newline
\begin{enumerate}[start=1,label={(\bfseries C\arabic*):}]
\item \begin{center} $x_{11}$ $<$ $x_{22}$ \end{center}
\item \begin{center} $x_{12}$ $>$ $x_{21}$ \end{center}
\item \begin{center} $y_{11}$ $<$ $y_{22}$ \end{center}
\item \begin{center} $y_{12}$ $>$ $y_{21}$ \end{center}
\end{enumerate}
\begin{figure}[!htbp]
  \centering
    \includegraphics[width=0.60\linewidth]{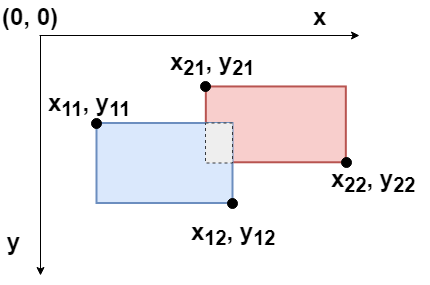}
  \caption{Slots are the isolated foregrounds represented as rectangle bounding boxes. Finding isolated bounding boxes can be achieved by analysing coordinates according to a series rules.}
\label{fig:system-architecture}
\end{figure}
The system pipeline contains two main phases: initialisation and image augmentation as shown in Figure \ref{fig:system-architecture}. 
To be specific, the valid slots selected by matching the complementary conditions proposed above. 
These picked slots are recorded into a database with detailed information include slot location, original foreground category and other scale related values (width, height, area size and aspect ratio). 
Generating images is conducted by substituting slots satisfying specific schemes determined by certain scenarios and scenarios are determined based the model performance evaluated by mAP and Derek P-R curve. 
For example, if the model behaves low mAP at a certain category, the augmentation is set to produce images containing objects of the target category, Figure \ref{fig:demo-same} gives a detailed workflow of augmenting extra images containing objects of the ``television" category. 
Many other scenario augmentation is presented in later sections of this chapter. 
\begin{figure}[!htbp]
  \centering
    \includegraphics[width=0.8\linewidth]{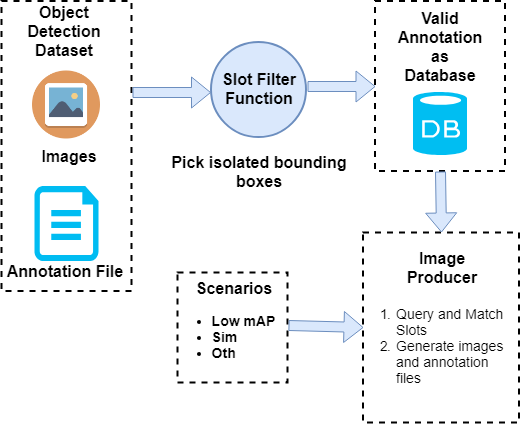}  
  \caption{Components of the Slot-based Image Augmentation System.}
\label{fig:system-architecture}
\end{figure}



%
In the reproduction and implementation aspect, the system consists several central steps such as system initialisation, performance analysis of detection model and image augmentation based on the decided scenario. 
In experiments, MS-COCO 2014 dataset is selected as the main dataset as it provides helpful API codes and contains a large amount of objects especially small objects \cite{lin2014microsoft}. 
MS-COCO 2014 is split as full training set, a subset of validation called "val minus minival" and the reduced validation dataset called "minival". 
The training set and randomly selected 5k "val minus minival" are used for traning while "minival" is used for validation. 
Furthermore, MS-COCO is a commonly used dataset in many related works as the reference of model capacity. 
For detection model selection, the PyTorch (\cite{paszke2017automatic}) version faster RCNN is selected as the benchmark detection model implemented by \cite{chen17implementation} which uses ImageNet pre-trained ResNet-50 as its backend proposed by \citep{imagenet_cvpr09,he2016deep}. 
Faster RCNN is a widely used model to compare with related works or transfer pre-trained weights and sufficiently supported by the deep learning community. 
\newline 

As in the pipeline Figure \ref{fig:system-architecture}, system initialisation follows loading dataset into the system, select isolated slots and summarise them into a database. 
In practice, dataset loading is implemented based on customised MS-COCO API to read annotation (JavaScript), load training images and refine the data structure. 
The associated Python libraries include Numpy, Scikit-Image, Matplotlib and many others \citep{van2011numpy,van2014scikit,hunter2007matplotlib}. 
The isolated slots selection is conducted by implementing the condition matching mentioned above, mainly by Numpy library. 
Database is implemented using data structure of Python list and dictionary encapsulated by PANDAS library (\cite{mckinney2011pandas}), which is feasible for matching SQL (Structured Query Language) style enquiry in a fast and efficient way especially handling large amount of data records. 
In addition, the slot aspect ratio, area size and other information are calculated using Numpy and Jupyter Notebook is the principle tool for efficient coding. 
\newline 

To analyse model performance with mAP, MS-COCO API provide evaluation tools in Python, Matlab and other languages. 
In practice, Python version official evaluation tools are integrated in the implementation of faster RCNN benchmark model. 
Since Derek P-R curve plotting is not supported by Python API, the Matlab version is used for the curve drawing and extra Linux shell scripts are implemented for the intermediate data structural transformation between Python and Matlab API. 
\newline 

Finally for the slot substitutions, the slot based sub-image cropping and pasting is conducted by python libraries of OpenCV and Scikit-Image \citep{opencv_library,van2014scikit}. 
Besides the crop and paste step, there are normally multiple available candidates for a certain slot and the rules of candidate selection is another important problem to concern.   
\section{Filtering Slot Candidates}
As mentioned in previous sections, slot-based image augmentation is conducted by substituting foreground objects and re-scale the candidate foreground to fit the slot shape. 
It is a common case that many candidate foregrounds are suitable for the target slot. 
Hence a candidate selection rule is required to address the ``many-to-one" issue. 
In this section, three different filters are introduced to ensure the augmentation quality with analysis and experiments presented. 
\newline 

In practice, an ``attribute matching" strategy is proposed to select valid candidates for the substitution with less damage to the potential learn-able features of the background image and candidate. 
Regarding the instance information in a large dataset, instances have a wide range of scale related attributes such as area size, length, width and height. 
Among those attributes, the instance area size, aspect ratio and category are applied as the filters to select candidates for slot substitution, in which the area size and aspect ratio are calculated in the format of rectangle bounding boxes. 
\newline 
\begin{figure}[!htbp]
  \centering
    \includegraphics[width=0.75\linewidth]{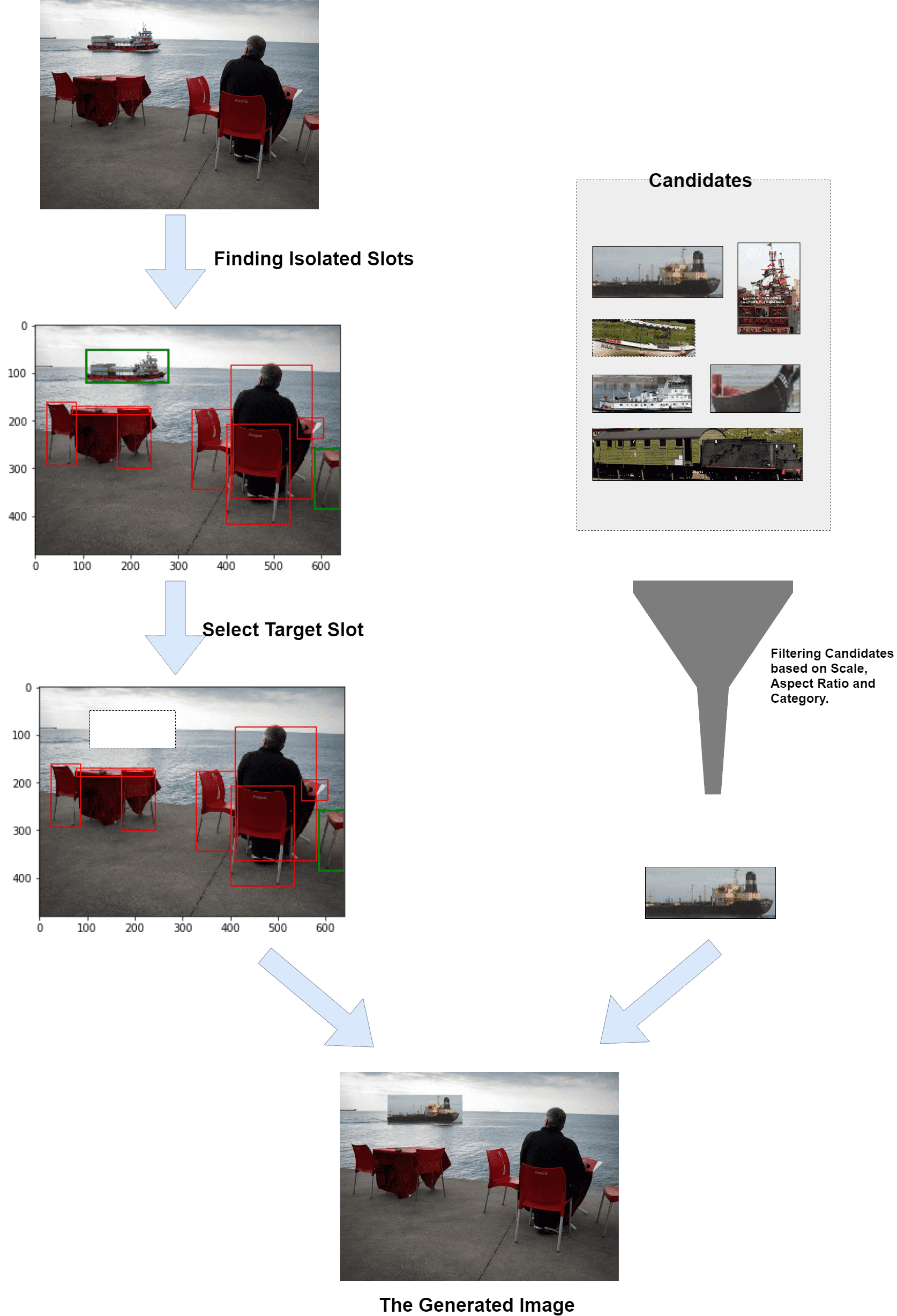}  
  \caption{System Work-flow}
\label{fig:system-work-flow}
\end{figure}
As the example in Figure \ref{fig:system-work-flow}, candidates are firstly filtered by their aspect ratio to avoid invalid re-scaling. 
Then candidates with similar scales/resolution are selected by the scale filter. 
Final step is filter rest of these candidates by their categories. 
The category filter is a scenario driven component. 
For example, candidates of the same category as the original slot foreground are selected when the augmentation system is aimed at augmenting extra images for a certain category. 
Another scenario is augmenting images to improve detection performance of one certain super-category instances such as substitute cats with dogs under animal super-category. 
Overall, filters are decided to preserve as much object information as possible without corrupting background images.
These three filters are simple, effective and easy for modification. 
Candidates selected by the three filters are randomly chosen as the final candidate to fit the target slot. 
\newline 

Furthermore, several experiments are conducted to test the validity of selecting these filters and a loss function based candidate matching policy is introduced to compare with the random selection method for the final candidate decision. 
To transfer the scenario in a more realistic level, a mini-COCO dataset is established for filter validation tests, presented in the following section.  
\subsection{Filter Validation on ``mini-COCO" dataset}
\begin{table}[h]
\label{table:inbalanced-dataset}
\caption{Top and Bottom Three Categories of Instance Amount in MS-COCO}
\begin{center}
\begin{tabular}{lll}
\hline
\multicolumn{1}{l|}{Categories}          & Image Amount    & Instance Amount  \\ \hline
\multicolumn{1}{l|}{\textbf{Person}}     & \textbf{64,115} & \textbf{262,465} \\
\multicolumn{1}{l|}{Chair}               & 12,774          & 38,491           \\
\multicolumn{1}{l|}{Car}                 & 12,251          & 43,867           \\ \hline
....                                     &                 &                  \\ \hline
\multicolumn{1}{l|}{Parking Meter}       & 705             & 1,285            \\
\multicolumn{1}{l|}{Toaster}             & 217             & 225              \\
\multicolumn{1}{l|}{\textbf{Hair Drier}} & \textbf{189}    & \textbf{198}     \\ \hline
\end{tabular}
\end{center}
\end{table}

As previously mentioned, MS-COCO dataset consists of images and instances, providing rich resource for slots and substitution candidates. 
However, lacking of data samples is a common issue due to the limited amounts of slots and candidates. 
To ensure the selected slot-based filters are functional in general conditions, it is necessary to conduct validation experiments on simulated scenarios. 
Hence, a mini-COCO dataset is established to run these tests with reduced image and instance amounts, limited slots and preserving the consistence with original MS-COCO. 
The mini-COCO dataset is required with the followings features: 
\newline 
\begin{enumerate}
    \item Reduced instance and image amounts to simulate the realistic scenario.
    \item All MS-COCO instance categories/super-category are included.
    \item Reasonable slot amount.
    \item Relatively balanced in an instance level. Table \ref{table:inbalanced-dataset} has listed some of the category amounts in MS-COCO, in which some categories have much more instances than the other.  
\end{enumerate}

\begin{figure}[!htbp]
\subfigure[]{\label{fig:img-num-epoch} \includegraphics[width=.40\textwidth]{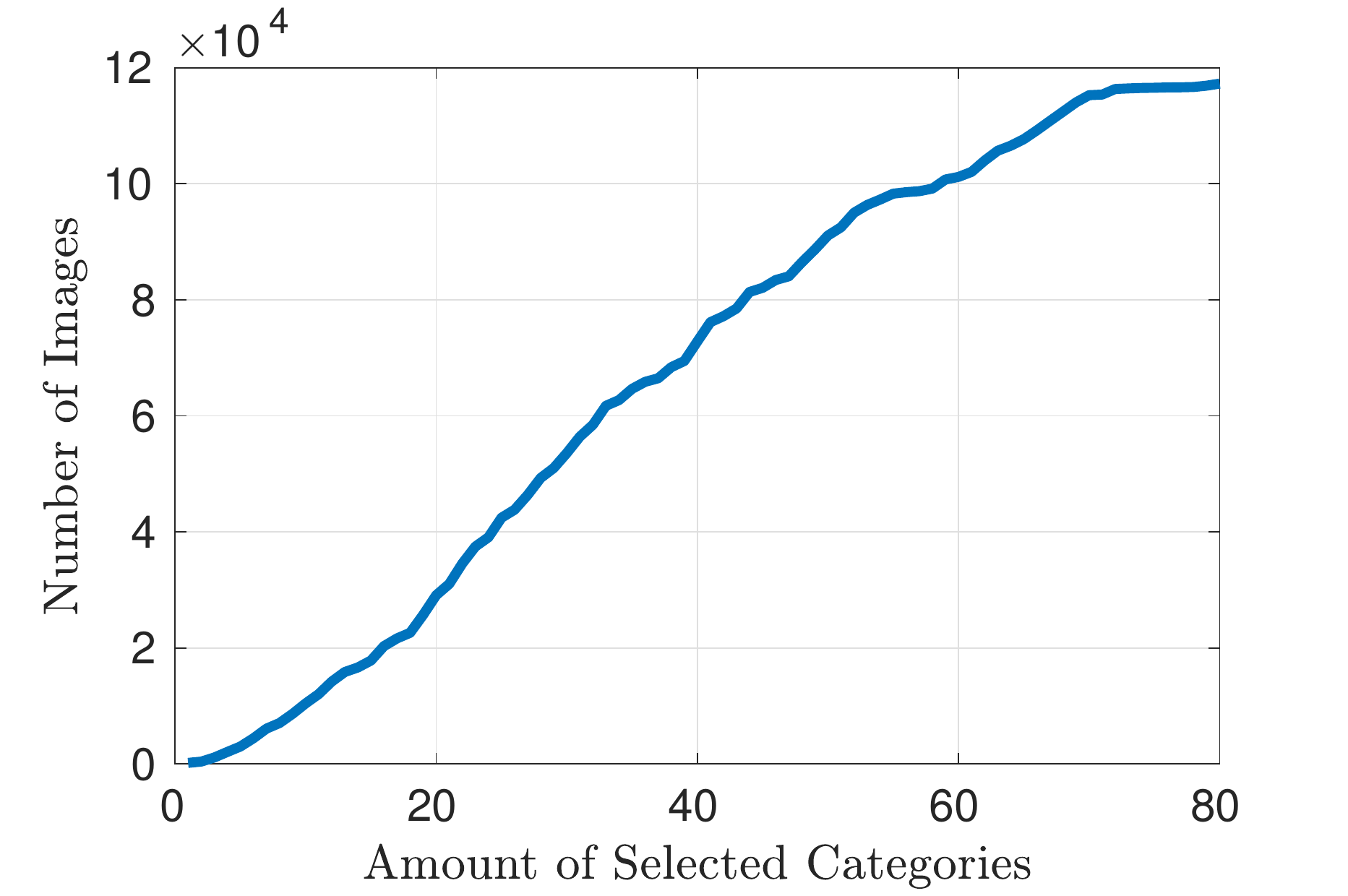}}
\hfill
\subfigure[]{\label{fig:inst-num-epoch} \includegraphics[width=.40\textwidth]{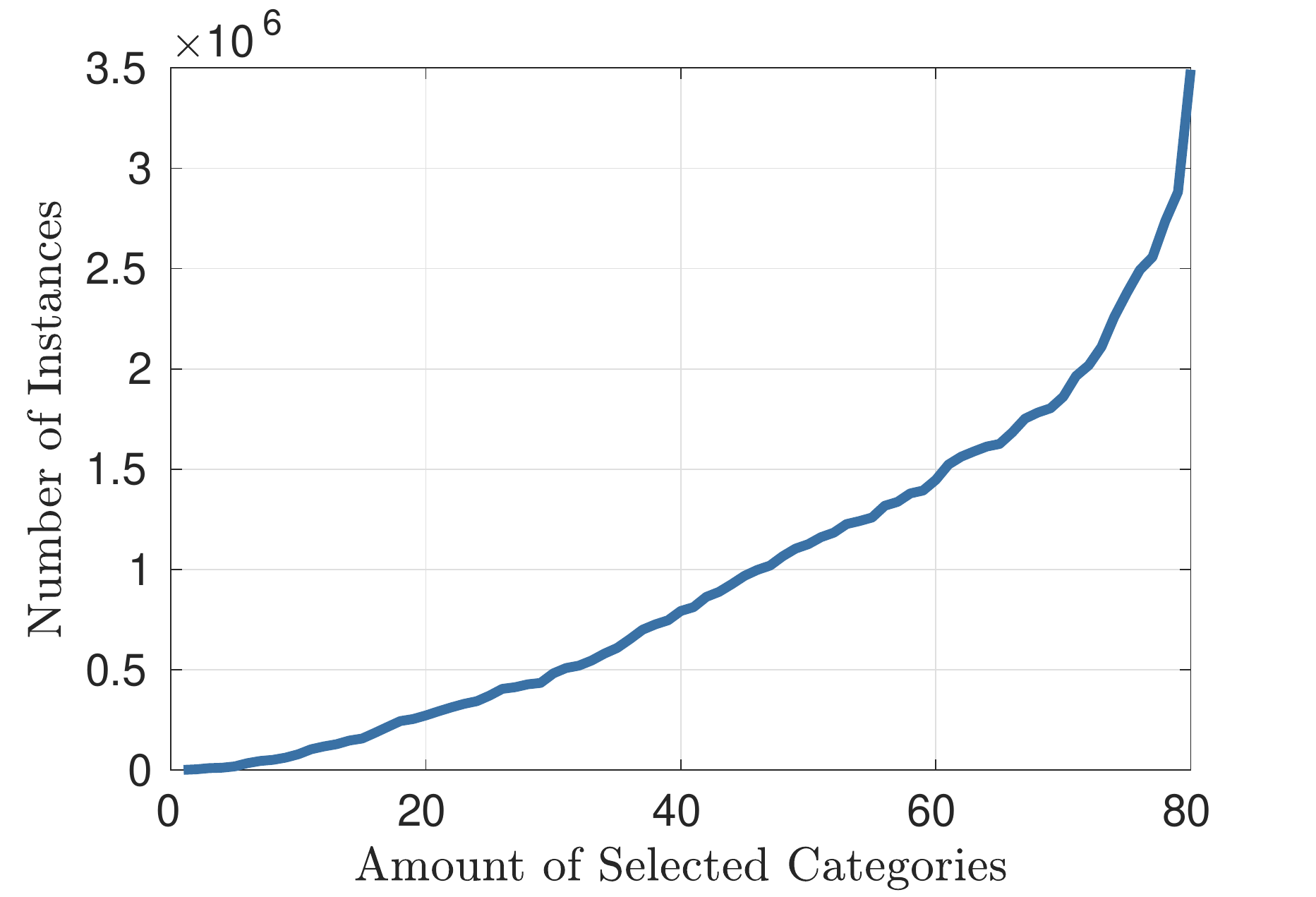}}
\hfill

\subfigure[]{\label{fig:num-slot-epoch.eps} \includegraphics[width=.40\textwidth]{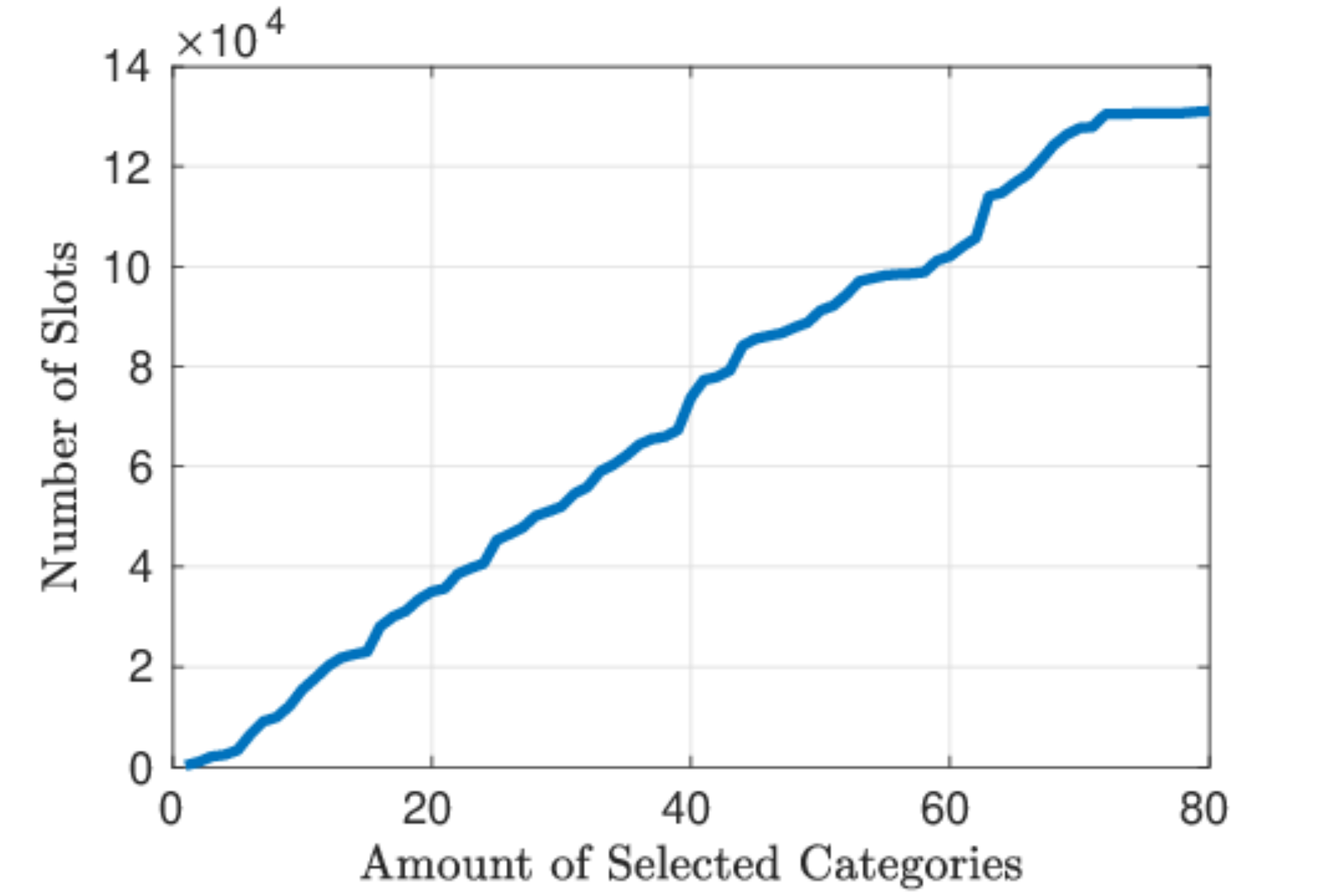}}
\hfill
\subfigure[]{\label{fig:avg-slot-epoch} \includegraphics[width=.40\textwidth]{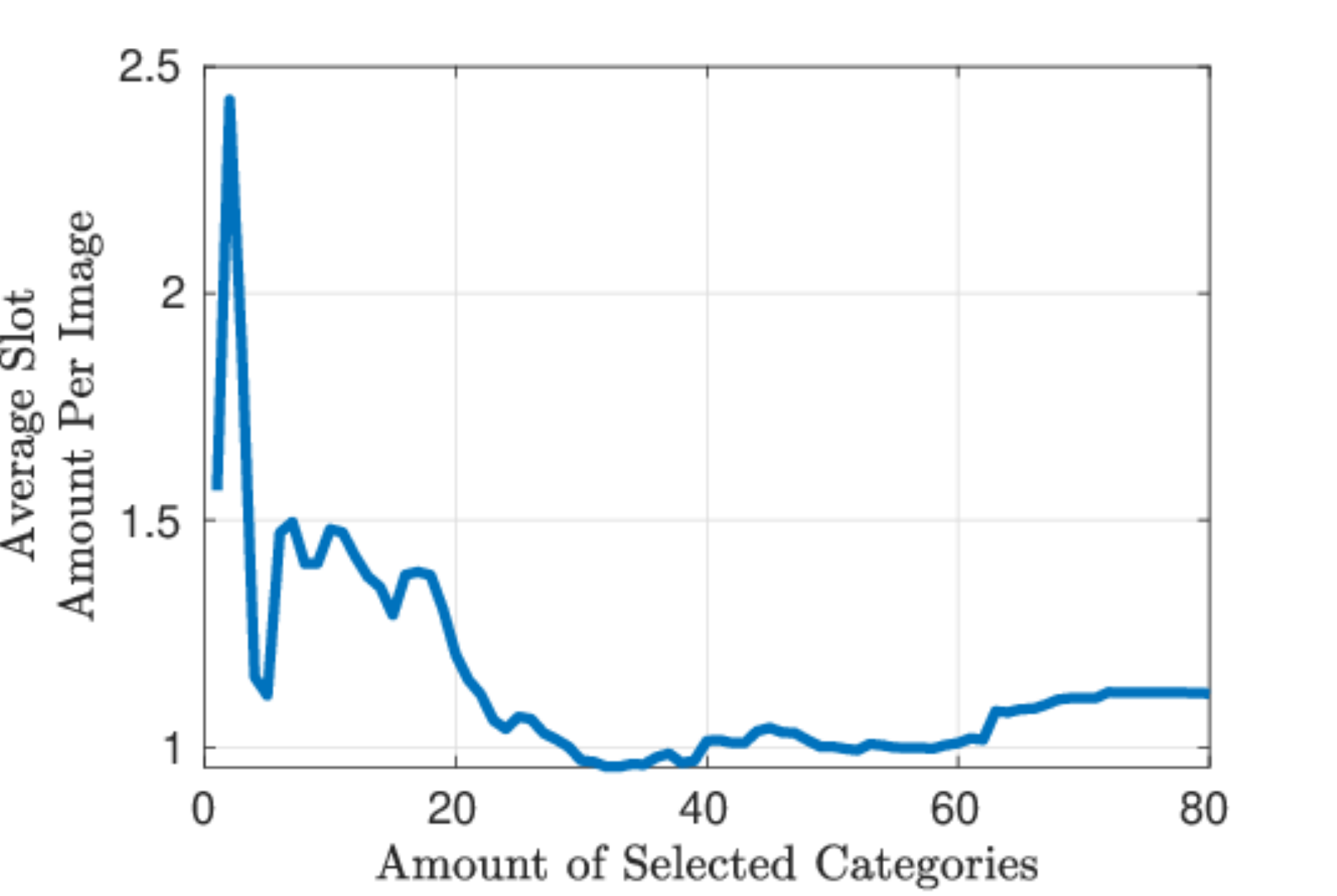}}
\hfill

\subfigure[]{\label{fig:std-epoch} \includegraphics[width=.40\textwidth]{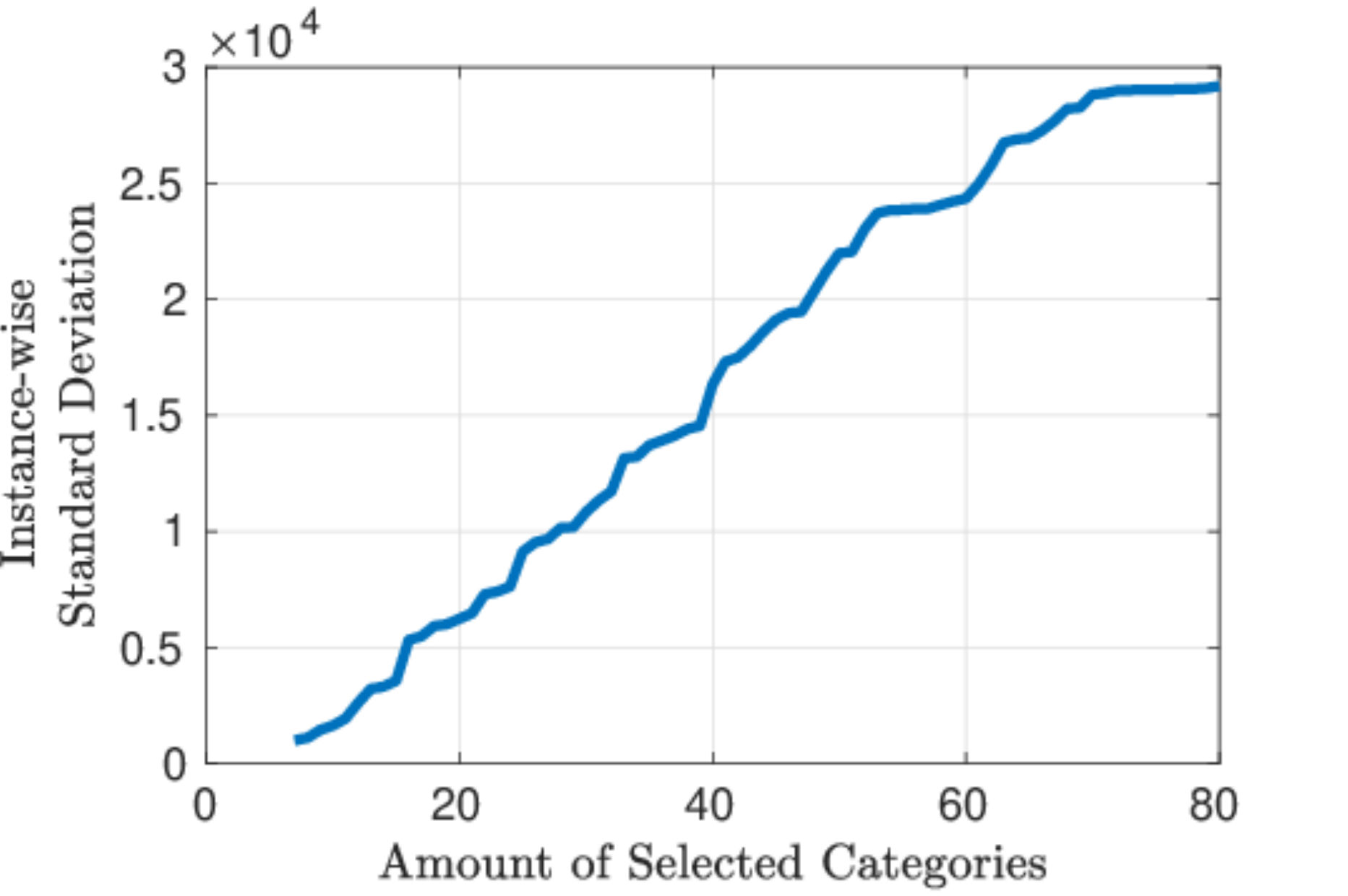}}
\hfill
\subfigure[]{\label{fig:capacity-epoch} \includegraphics[width=.40\textwidth]{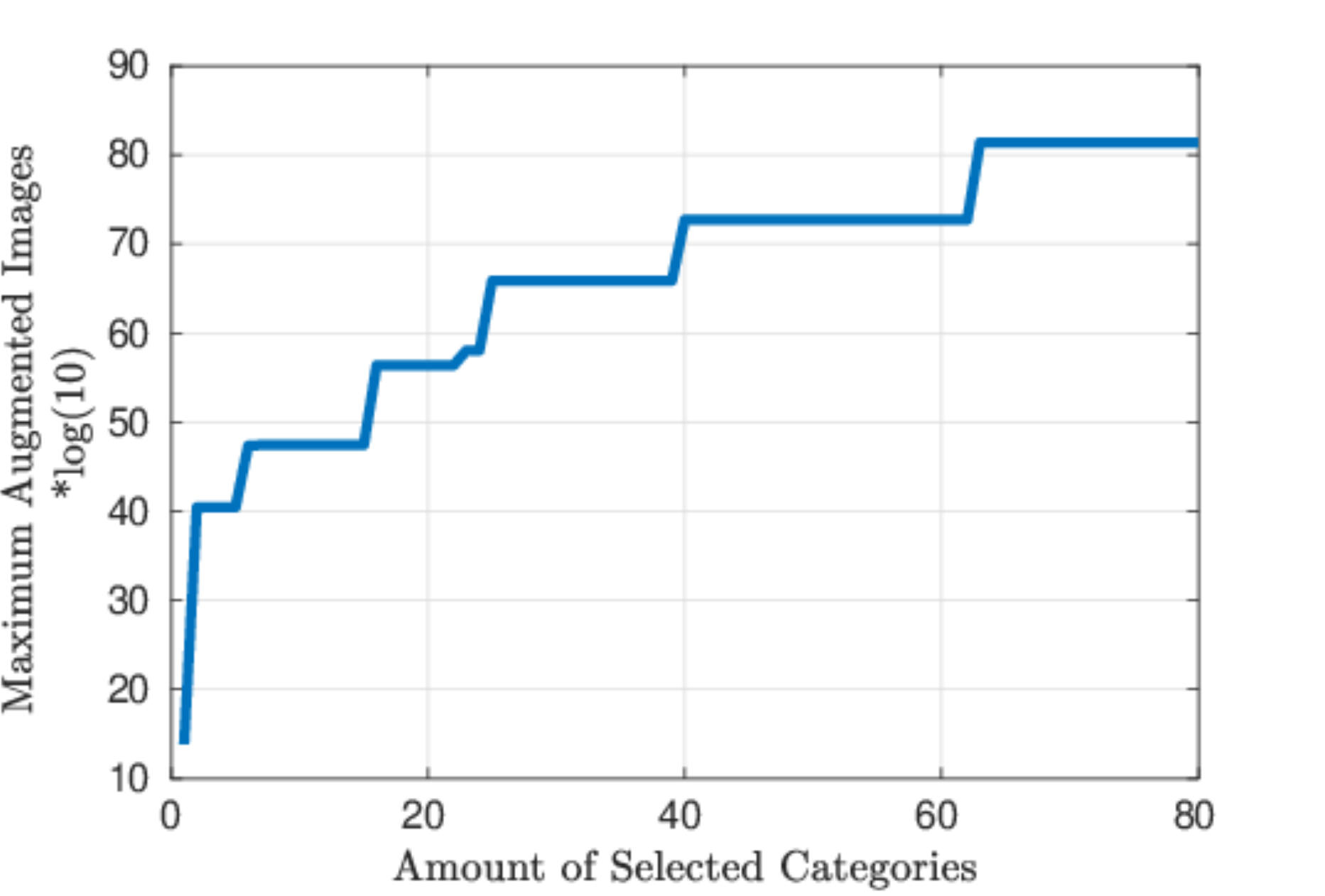}}
\caption{\textbf{Row one:} image and instance amount as selecting more categories. \textbf{Row two:} overall slot amount and average slot number per image of mini-coco as selecting more categories. \textbf{Row three:} instance standard deviation and augmentation capacity.}
\label{fig:metrics-num-epoch}
\end{figure}

To maintain those desired features, mini-COCO dataset is established in a heuristic way to iteratively fetch images from MS-COCO as presented in Algorithm \ref{alg:construct-minicoco}. 
Based on MS-COCO annotations, all 80 categories are sorted in an ascending order of instance amount. 
After some other data structure wise setup, the desired mini-coco dataset is selected from a set of candidates delivered by an accumulative operation of stacking images containing a certain category instances. 
In other words, a single category with the smallest instance amount is selected by previous sorting operation and those images containing instances of the selected category, are added into mini-coco as a candidate. 
As the iterative progress, these mini-coco candidates are evaluated with a series of metrics including standard deviation of instance amounts for dataset balancing concern, slot amount, summary of instance and image amounts and whether all categories are included. 
The final mini-coco is selected based on the comparison of those feature-related metrics and in our case, mini-coco dataset is selected at the 20th category is selected containing 73,653 instances in 10,436 images with all instance categories included. 
In addition, mini-coco contains approximately 9\% images (8.8\%) and instances (8.6\%) comparing with the original MS-COCO which contains 118,287 images and 860,001 instances. 
Figure \ref{fig:metrics-num-epoch} has presented the related metric values as stacking categories into the mini-coco candidates.


\begin{algorithm}
\label{alg:construct-minicoco}
\caption{A heuristic algorithm to construct mini-COCO dataset}\label{alg:construct mini-coco}
\begin{algorithmic}[1]
\Procedure{Construct Mini-COCO}{}
\State Initialisation:
\State $num\_cat$ \Comment{category amount in MS-COCO}
\State $sorted\_array\ \gets$ MS-COCO categories sorted by instance amount in an ascending order
\State $mini\_coco\ \gets\ []$
\State $records\ \gets\ []$ \Comment{calculate and store values corresponding to dataset requirements including standard deviation (std), slot amount (slot\_amount) and check whether all categories are included (category\_included)}
\State Progress 
\While{$i$ from $0$ to $num\_cat-1$}\Comment{iterate items in $sorted\_array$}
\State $mini\_coco[i]\ \gets$ images and associated annotations containing category $sorted\_array[i]$
\State CLEANUP \Comment{remove selected images and annotations from MS-COCO}
\State $records[i]\ \gets$ $std$, $slot\_amount$ and $category_included$
\EndWhile
\State Output $records$ and $mini\_coco$
\State \textbf{RETURN} $mini\_coco[a]$ \Comment $mini\_coco[a]$ is manually chosen based on $records$ and $mini\_coco$.
\EndProcedure
\end{algorithmic}
\end{algorithm}

\subsection{Filter Details and Validation}
As discussed above, the chosen filters are aspect ratio, scale and category. 
In this section experiments are conducted on testing turning on and off these filters using mini-coco dataset and baseline faster RCNN model. 
\newline 

In practice, filters are defined as a selector to pick valid candidates from a certain range w.r.t. the scheme. 
For example, the scale filter compares candidates' scales and preserve those within $\pm 20\%$ of the original scales. 
The aspect ratio filter works in the same way and the category filter. 
The proportion threshold $\pm 20\%$ is a trade off between candidate quality and candidate amount. 
The threshold is preferred to be larger When the dataset includes large number of instances and smaller with relatively small dataset. 
Category filter is applied according to specific scenario. 
For the task of generating more images with same category, candidates are preserved only if they belong to the same category as the original slot instances. 
\newline 

\begin{figure}[!htbp]
    \centering
    \includegraphics[width=.4\linewidth]{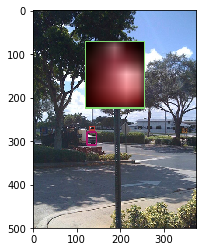}
    \caption{A negatively generated image, in which the street sign is too small to fit into the slot.}
    \label{fig:scale-matters}
\end{figure}

\begin{figure}[!htbp]
    \centering
    \includegraphics[width=.4\linewidth]{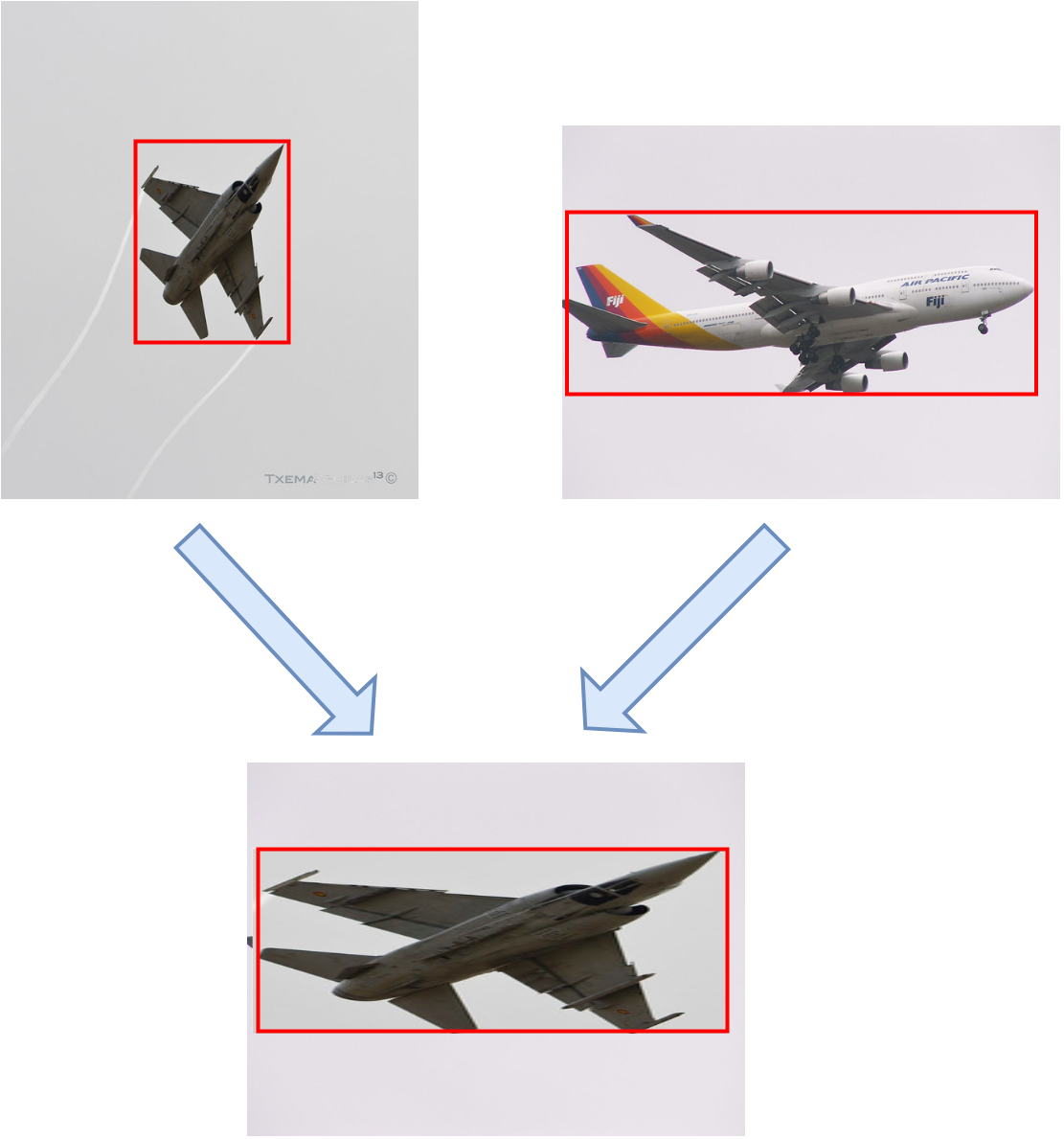}
    \caption{Object feature loss is caused by substituting the target slot with an instance of different aspect ratio. In this case, the target slot ratio is width larger than height while the substitution is opposite.}
    \label{fig:ar-matters}
\end{figure}

\begin{table}[]
\label{table:filter-on-off}
\caption{Filter ON/OFF Validations}
\begin{center}
\begin{tabular}{@{}ccc|c@{}}
\toprule
\textbf{Category} & \textbf{Scale} & \textbf{Aspect Ratio} & \textbf{mAP}   \\ \midrule
\textbf{ON}                & \textbf{ON}             & \textbf{ON}                    & \textbf{0.149} \\
OFF               & ON             & ON                    & 0.146          \\
ON                & ON             & OFF                   & 0.147          \\
ON                & OFF            & ON                    & 0.148          \\ \bottomrule
\end{tabular}
\end{center}
\end{table}
Figure \ref{table:filter-on-off} shows the experiments on baseline model testing turning on or off those filters on mini-coco dataset. 
The experimental results demonstrates the necessary of turning on these filters and Figure \ref{fig:scale-matters} gives an invalid example image without turning on the scale filter and Figure \ref{fig:ar-matters} presents the case of inappropriate aspect ratio. 

\section{Performance-based Image Augmentation}
With the aforementioned filters, slot-based augmentation system can be applied in many different scenarios and this section presents an example of improving detection performance by generating images with the substitution of same category instances. 
\newline 
\begin{figure}[!htbp]
\subfigure[overall mAP]{\label{fig:nofilp-overall-all} \includegraphics[width=.40\textwidth]{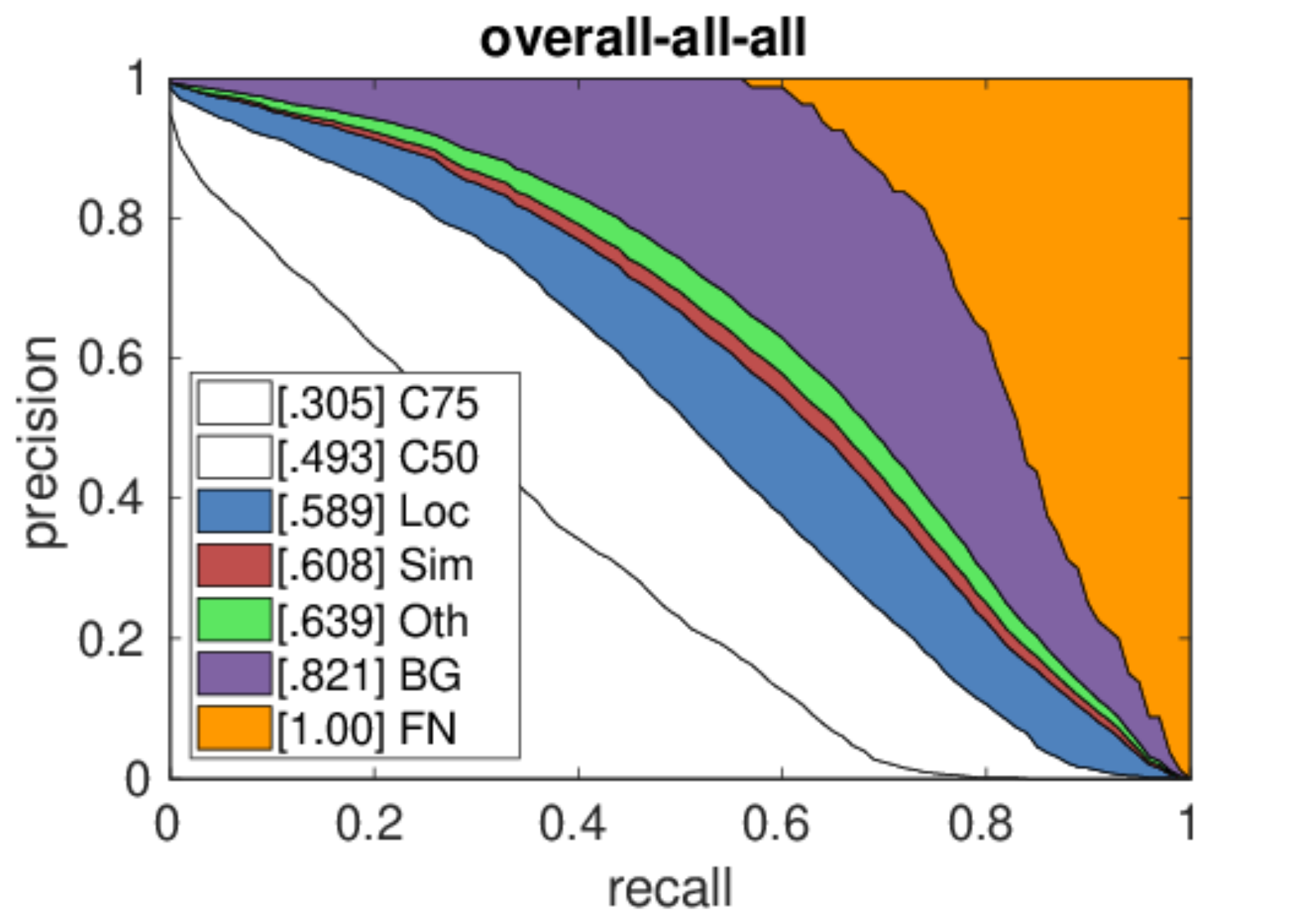}}
\hfill
\subfigure[]{\label{fig:nofilp-overall-large} \includegraphics[width=.40\textwidth]{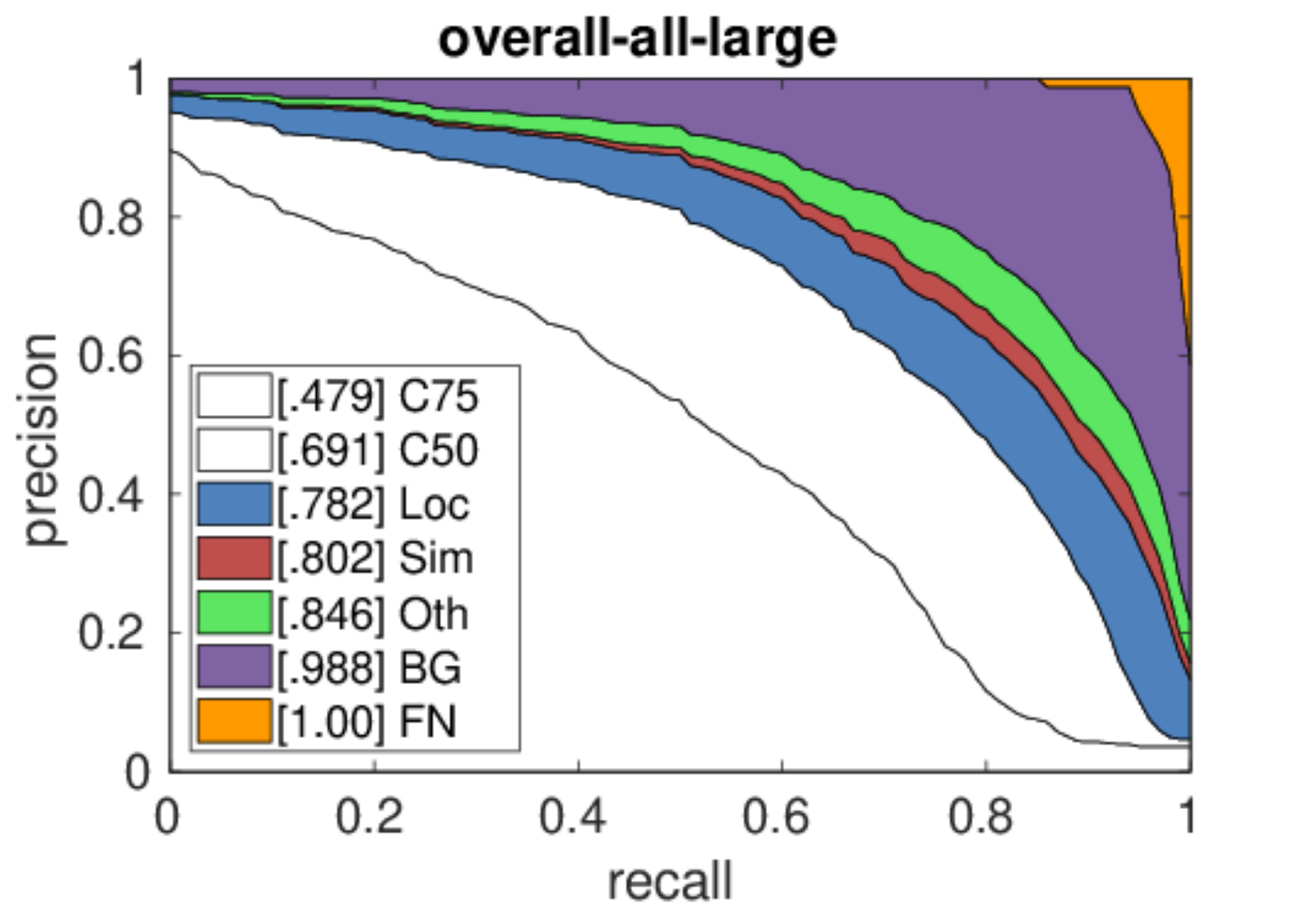}}
\hfill

\subfigure[]{\label{fig:nofilp-overall-medium} \includegraphics[width=.40\textwidth]{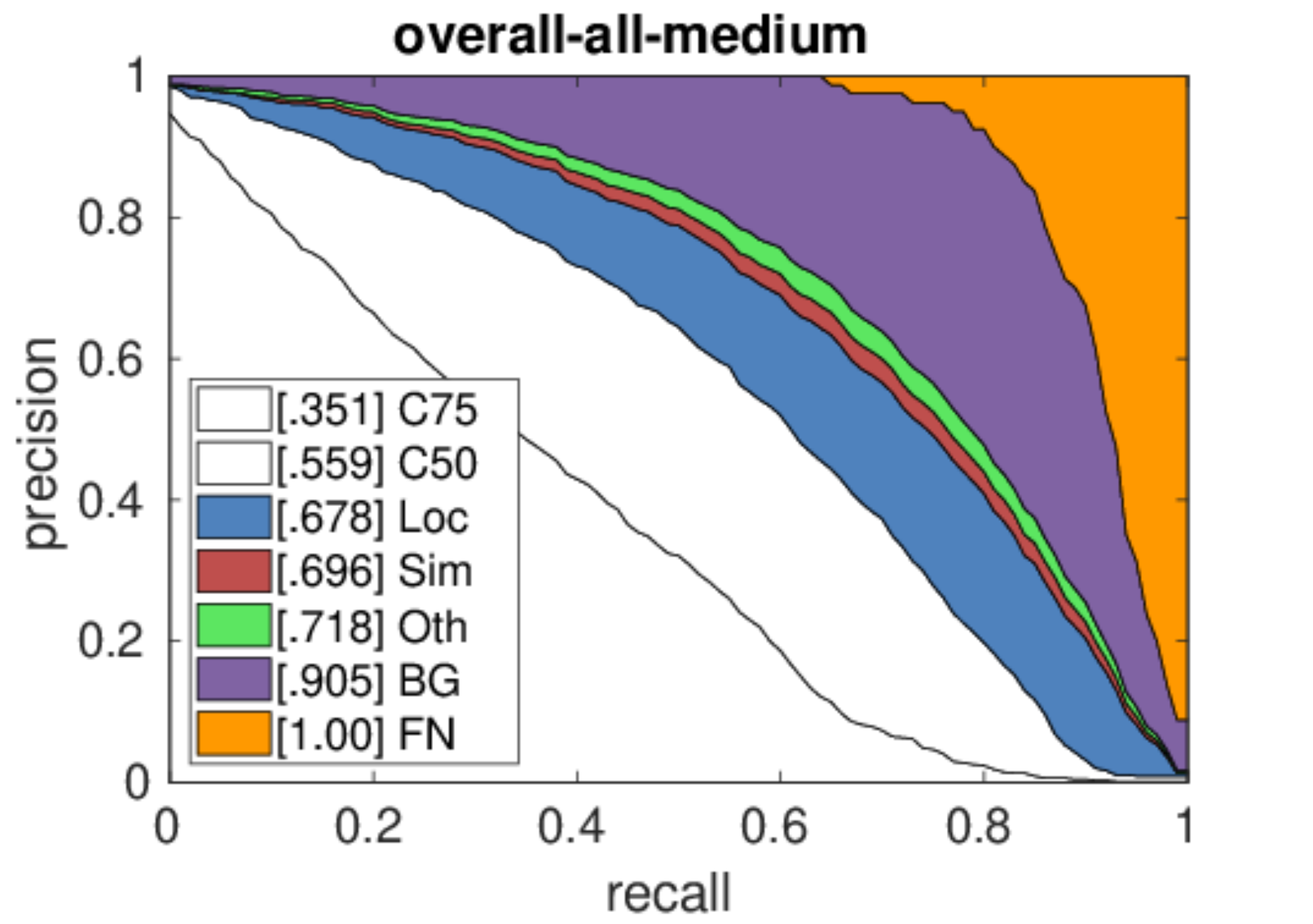}}
\hfill
\subfigure[]{\label{fig:nofilp-overall-small} \includegraphics[width=.40\textwidth]{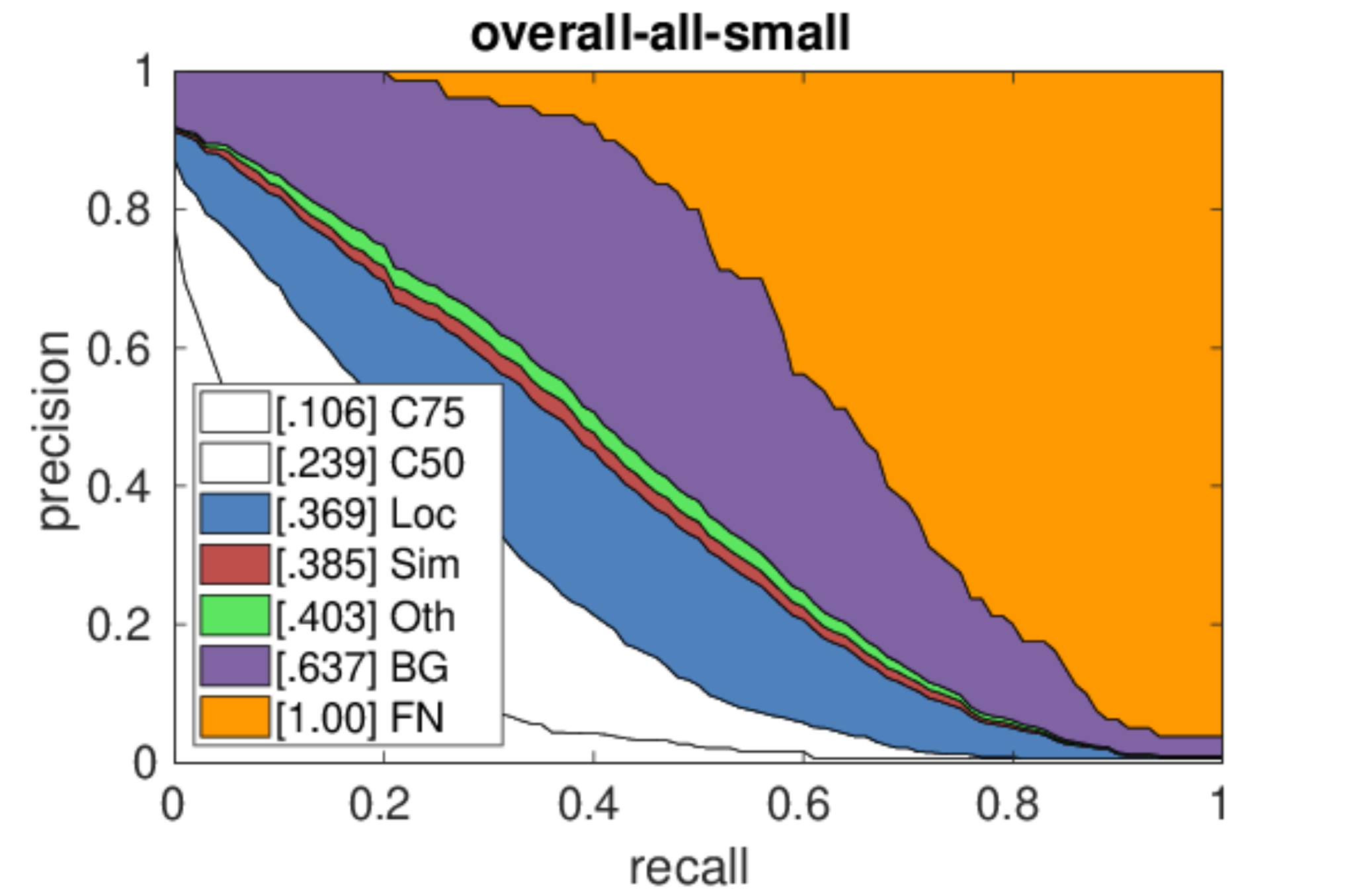}}
\caption{Derek Style P-R Curves of Baseline Faster RCNN Model without Image Augmentation.}
\label{fig:noflip-overall}
\end{figure}
The performance-based image augmentation is based on the training mAP and P-R details of the baseline model. 
In this case, the original faster RCNN model is selected as baseline model and trained on MS-COCO dataset with 490k batches without any augmentation (same hyperparameter configuration as \cite{fasterrcnn}). 
The baseline model produces 30.5\% overall mAP (C75) that reaches the original benchmark. 
However, mAP alone is biased to describe model performance and weakness. 
Hence, detailed P-R curves in Figure \ref{fig:noflip-overall} are introduced to evaluate the scale-wise performance. 
In addition to the overall performance, category-wise P-R curves are analysed as well. 
The P-R curve and detailed mAP performance indicates the imbalanced performance that some categories have fairly low mAP due to limited image and instance amount. 
To address the issue of low mAP of certain categories, augmenting extra images to provide extra learn-able features is expected a performance improvement, hence the motivation of using slot-based augmentation system. 
\newline 

Based on the analysis of the mAP and instance amount, three categories are selected to demonstrate the process of augmenting images to improve the performance: \textbf{cars}, \textbf{ships} and \textbf{traffic lights}. 
These three categories initially have less instance amounts, lower detection mAP and reasonable number of slot. 
To augment more images containing these category instances, the slots are substituted with candidates of the same category and the $\pm 20\%$ threshold is applied on the scale and aspect ratio filters. 
Based on the selected categories, the widely used flipping and slot-based approaches are compared in the aspects of various mAP metrics, the amount of generated images and the effects of combining them together. 
Table \ref{table:augment-amount} displays the augmentation amount with different augmentation methods, in which flipping images is the commonly used approach in many detection systems. 
In addition, these augmented images also change the probability of training on an original image in other words, the chances of learning original features or artificial features. 
Details of the augmentation performance are presented in the following sections. 
\textbf{Note} the slot-based method is conducted one epoch so that one slot is only substituted with one candidate. 

\begin{table}[]
\label{table:augment-amount}
\caption{Number of Augmented Images with Different Approaches}
\begin{center}
\begin{tabular}{@{}l|l|l|l@{}}
\hline
Augmentation Methods(490k)       & \begin{tabular}[c]{@{}l@{}}Number of \\ Original Images\end{tabular} & \begin{tabular}[c]{@{}l@{}}Number of \\ Augmented Image\end{tabular} & \begin{tabular}[c]{@{}l@{}}Proportion of \\ Original Images\end{tabular} \\ \hline
No-Flipping                      & 118,287                       & 0                     & 0\%                                        \\
Flipping (All Categories)                         & 118,287                       & 118,287               & 50\%                                       \\ \midrule
Ours Cars                        & 12,251                   & 3,262                  & 78.97\%                                         \\
Ours Boats                       & 3,025                  & 940                   & 76.29\%                                         \\
Ours Traffic Lights              & 4,139         & 2,224                  & 64.05\%                                         \\ \midrule
Flipping and Ours Cars           & 12,251                   & 15,513            & 44.13\%                                         \\
Flipping and Ours Boats          & 3,025                  & 3,956            & 43.31\%                                         \\
Flipping and Ours Traffic Lights & 4,139         & 6,363           & 39.41\%                                         \\ \bottomrule
\end{tabular}
\end{center}

\end{table}

\subsection{Augmenting Cars Images}
MS-COCO contains 43,867 car instances distributed in 12,251 images. 
Following the pipeline discussed in Section \ref{subsection:augmentation-pipeline}, isolated slots are selected and replaced with candidates satisfying the requirements of the filters. 
In other words, the car slots are substituted with those candidates with close aspect ratio, scales and the same category. 
In the experiments, 3,262 images are generated by replacing slots by one iteration and the sample images are shown in Figure \ref{fig:sub-first}. 
\newline 

Regarding augmentation effects on detection performance, Table \ref{table:cars-overall} presents baseline model mAP with different augmentation methods of all scales. 
Table \ref{table:cars-large}, Table \ref{table:cars-medium} and Table \ref{table:cars-small} describe the same schemes on objects with scales of large, medium and small relatively. 
To summarise these results, slot-based augmentation system generates around 3000 images (over 12,251 images) and train on these extra images improved the overall mAP 2\% comparing with commonly used ``\textbf{flipping}" and non-augmentation. 
Regarding the training mAP of different scales, it can observed that slot-based augmentation has greater effects for medium and small objects, boosting 1.6\% mAP for medium instances and 0.9\% for small scale (comparing with not applying any augmentation methods, under the \textbf{C75} metric).  
Furthermore, the mixing of slot-based method and flipping, increased 1.4\% mAP while 0.5\% for flipping only. 
Despite the mAP improvements of flipping, the overall mAP from Table \ref{table:cars-overall} shows a 0.7\% mAP decreasing which might be the reason of medium scale instances in Table \ref{table:cars-medium}.  

\begin{figure}[!htbp]
\subfigure[]{\label{fig:sub-first} \includegraphics[width=.25\textwidth]{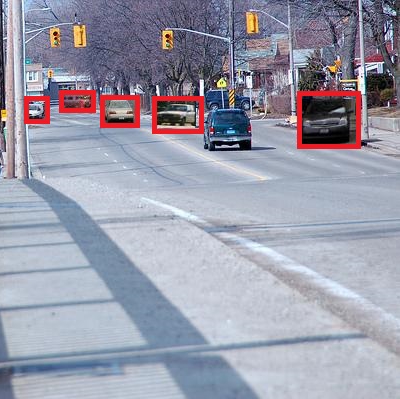}}
\hfill
\subfigure[]{\label{fig:sub-first} \includegraphics[width=.25\textwidth]{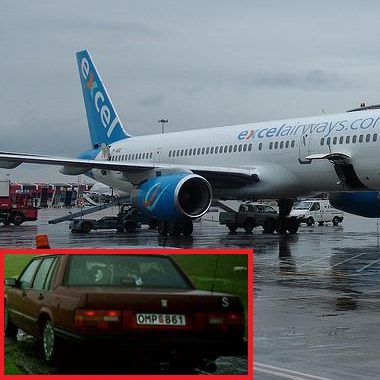}}
\hfill
\subfigure[]{\label{fig:sub-first} \includegraphics[width=.25\textwidth]{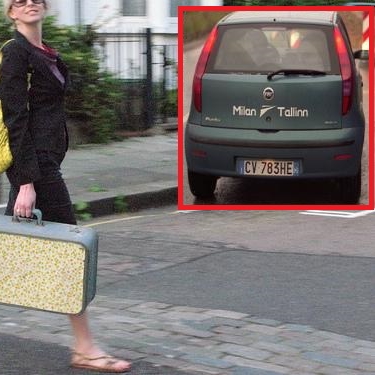}}
\hfill

\subfigure[]{\label{fig:sub-first} \includegraphics[width=.25\textwidth]{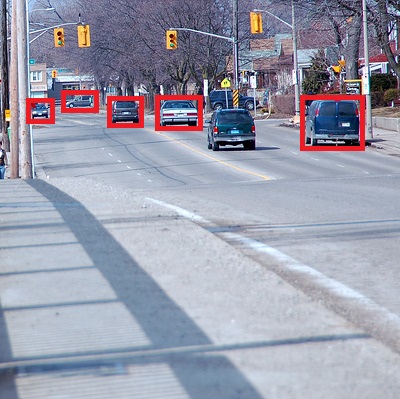}}
\hfill
\subfigure[]{\label{fig:sub-first} \includegraphics[width=.25\textwidth]{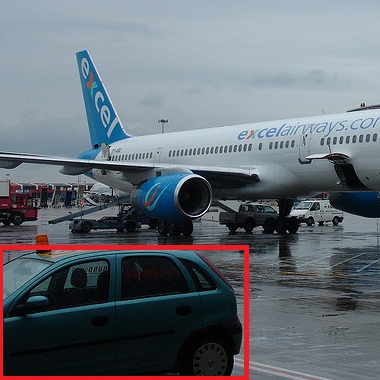}}
\hfill
\subfigure[]{\label{fig:sub-first} \includegraphics[width=.25\textwidth]{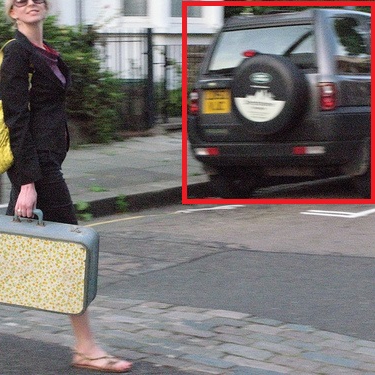}}
\caption{\textbf{Row one:} augmented car images. \textbf{Row two:} the original images (some images are cropped for presentation purpose).}
\label{fig:car-examples}
\end{figure}

\begin{table}[!htbp]
\label{table:cars-overall}
\caption{Baseline Model mAP on \textbf{Cars}}
\centering
\begin{tabular}{l|l|l|l}
                & C75   & C50   & Loc (C10) \\ \hline
No Flipping     & 0.268 & 0.513 & 0.688     \\
Flipping        & 0.261 & 0.510 & 0.691     \\
\textbf{Ours}            & \textbf{0.281} & \textbf{0.509} & \textbf{0.679}     \\
\textbf{Flipping + Ours} & \textbf{0.276} & \textbf{0.515} & \textbf{0.677}    
\end{tabular}
\end{table}
\begin{table}[!htbp]
\label{table:cars-large}
\centering
\caption{Baseline Model mAP on \textbf{Cars (Large Scale)}}
\begin{tabular}{l|l|l|l}
                & C75   & C50   & Loc (C10) \\ \hline
No Flipping     & 0.538 & 0.776 & 0.874     \\
Flipping        & 0.599 & 0.768 & 0.872     \\
\textbf{Ours}            & \textbf{0.580} & \textbf{0.778} & \textbf{0.883}     \\
\textbf{Flipping + Ours} & \textbf{0.538} & \textbf{0.802} & \textbf{0.883}    
\end{tabular}
\end{table}
\begin{table}[!htbp]
\label{table:cars-medium}
\centering
\caption{Baseline Model mAP on \textbf{Cars (Medium Scale)}}
\begin{tabular}{l|l|l|l}
                & C75            & C50            & Loc (C10)      \\ \hline
No Flipping     & 0.497          & 0.727          & 0.860          \\
Flipping        & 0.486          & 0.727          & 0.842          \\
\textbf{Ours }           & \textbf{0.513} & \textbf{0.736} & \textbf{0.848} \\
\textbf{Flipping + Ours} & \textbf{0.501}          & \textbf{0.729}          & \textbf{0.841}
\end{tabular}
\end{table}

\begin{table}[!htbp]
\label{table:cars-small}
\centering
\caption{Baseline Model mAP on \textbf{Cars (Small Scale)}}
\begin{tabular}{l|l|l|l}
                         & C75            & C50            & Loc (C10)      \\ \hline
No Flipping              & 0.118          & 0.384          & 0.620          \\
Flipping                 & 0.123          & 0.372          & 0.622          \\
\textbf{Ours}            & \textbf{0.127}          & \textbf{0.374 }         & \textbf{0.611}          \\
\textbf{Flipping + Ours} & \textbf{0.132} & \textbf{0.376} & \textbf{0.599}
\end{tabular}
\end{table}

\subsection{Augmenting Boats Images}
\begin{figure}[!htbp]
\subfigure[]{\label{fig:sub-first}\includegraphics[width=.25\textwidth]{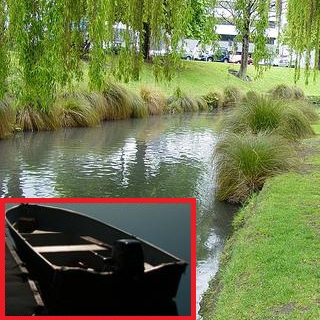}}
\hfill
\subfigure[]{\label{fig:sub-second}\includegraphics[width=.25\textwidth]{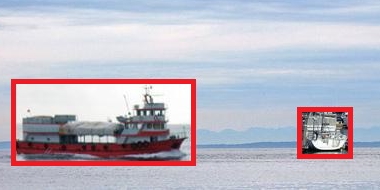}}
\hfill
\subfigure[]{\label{fig:sub-third}\includegraphics[width=.25\textwidth]{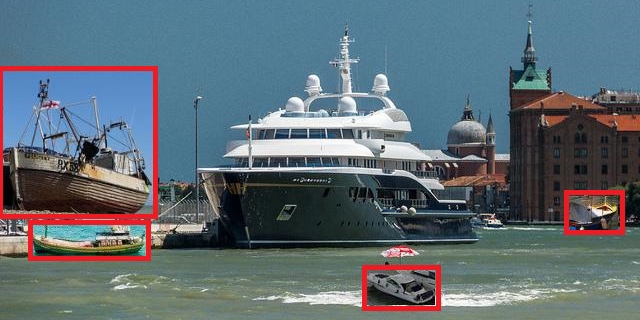}}
\hfill

\subfigure[]{\label{fig:sub-fourth} \includegraphics[width=.25\textwidth]{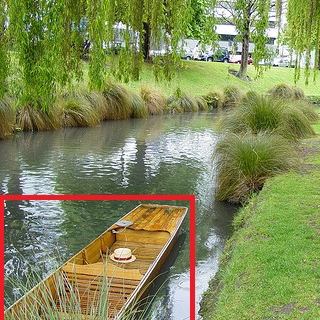}}
\hfill
\subfigure[]{\label{fig:sub-fifth} \includegraphics[width=.25\textwidth]{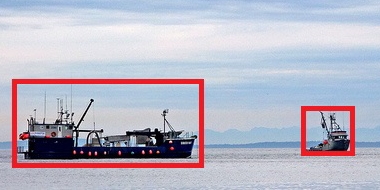}}
\hfill
\subfigure[]{\label{fig:sub-sixth} \includegraphics[width=.25\textwidth]{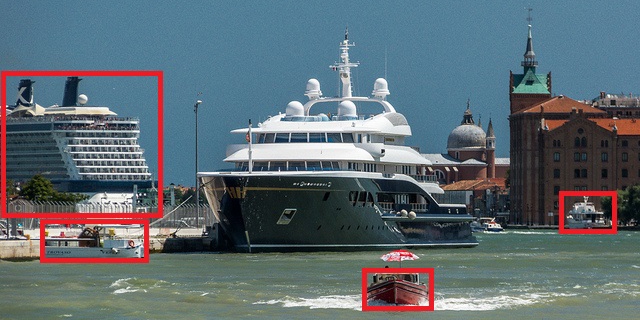}}

\caption{\textbf{Row one:} augmented boat images \textbf{Row two:} the original images}
\label{fig:boat-examples}
\end{figure}

Similar to augment images with car instances, after one iteration, 940 boat images are generated from 3,025 images with 10,759 instances. 
Figure \ref{fig:boat-examples} shows three sample images before and after slot substitution. 
Considering the training mAP, the flipping and slot-based augmentation methods actually decreased the mAP by 0.9 \% and 3.1\% respectively. 
While combining these two augmentation methods yields an increasing 2.3\% mAP. 
For different scale objects, the two augmentation methods decreased detection mAP at some extent. 
But slot-based method contributes a 2.8\% improvement for medium scale objects and flipping method rises a 1.3\% improvement on small scale objects. 
In general, for the boat-related image augmentation, individual augmentation failed to improve training mAP with a decreasing effect instead. 
However, applying these two methods together improved detection mAP at all scales. 


\begin{table}[!htbp]
\label{table:boats-overall}
\caption{Baseline Model mAP on \textbf{Boats}}
\centering
\begin{tabular}{l|l|l|l}
                         & C75            & C50            & Loc (C10)      \\ \hline
No Flipping              & 0.127          & 0.37           & 0.594          \\
Flipping                 & 0.118          & 0.366          & 0.600          \\
\textbf{Ours}            & \textbf{0.096}          & \textbf{0.351}          & \textbf{0.592}          \\
\textbf{Flipping + Ours} & \textbf{0.140} & \textbf{0.385} & \textbf{0.603}
\end{tabular}
\end{table}

\begin{table}[!htbp]
\label{table:boats-large}
\caption{Baseline Model mAP on \textbf{Boats (Large Scale)}}
\centering
\begin{tabular}{l|l|l|l}
                         & C75            & C50            & Loc (C10)      \\ \hline
No Flipping              & 0.342          & 0.639          & 0.778          \\
Flipping                 & 0.323          & 0.591          & 0.769          \\
\textbf{Ours}            & \textbf{0.239}          & \textbf{0.570 }         & \textbf{0.784}          \\
\textbf{Flipping + Ours} & \textbf{0.428} & \textbf{0.630} & \textbf{0.795}
\end{tabular}
\end{table}
\begin{table}[!htbp]
\label{table:boats-medium}
\caption{Baseline Model mAP on \textbf{Boats (Medium Scale)}}
\centering
\begin{tabular}{l|l|l|l}
                         & C75            & C50            & Loc (C10)      \\ \hline
No Flipping              & 0.125          & 0.474          & 0.727          \\
Flipping                 & 0.103          & 0.460          & 0.728          \\
\textbf{Ours}            & \textbf{0.153}          & \textbf{0.504}          & \textbf{0.757}          \\
\textbf{Flipping + Ours} & \textbf{0.134} & \textbf{0.428} & \textbf{0.711}
\end{tabular}
\end{table}
\begin{table}[!htbp]
\label{table:boats-small}
\caption{Baseline Model mAP on \textbf{Boats (Small Scale)}}
\centering
\begin{tabular}{l|l|l|l}
                         & C75            & C50            & Loc (C10)      \\ \hline
No Flipping              & 0.043          & 0.259          & 0.564          \\
Flipping                 & 0.056          & 0.272          & 0.585          \\
\textbf{Ours}            & \textbf{0.029}          & \textbf{0.230}          & \textbf{0.563}          \\
\textbf{Flipping + Ours} & \textbf{0.063} & \textbf{0.275} & \textbf{0.565}
\end{tabular}
\end{table}

\subsection{Augmenting Traffic Light Images}
For the traffic light category, 2,224 images are augmented based on 12,884 instances from 4,139 original images within one epoch. 
Some sample images are shown in Figure \ref{fig:sub-first}. 
In the aspect of detection mAP, both flipping and slot-based augmentation have improved the mAP. 
The improvement for flipping is 0.3\% and 2.0\% for slot-based method. 
The combination of these two method is 2.0\% as well. 
Observing the mAP changes on different scales, flipping method improved 3.2\% and 4.9\% for large and medium objects while a decreased 0.4\% for small objects. 
The slot-based method produces 2.9\% and 3\% for small objects. 
However, it decreased 0.7\% for medium objects. 
The combination approach improved all the three scales with 3.5\%, 3.3\% and 1.5\% for large, medium and small objects respectively. 
\begin{figure}[!htbp]
\subfigure[]{\label{fig:sub-first}\includegraphics[width=.25\textwidth]{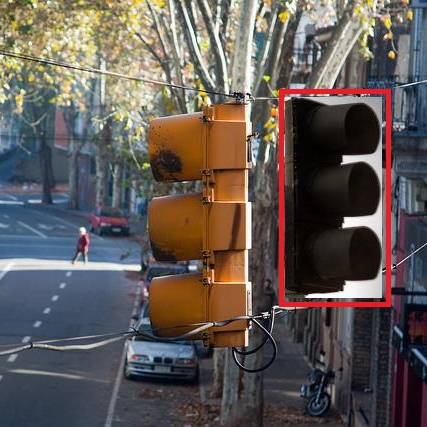}}
\hfill
\subfigure[]{\label{fig:sub-second}\includegraphics[width=.25\textwidth]{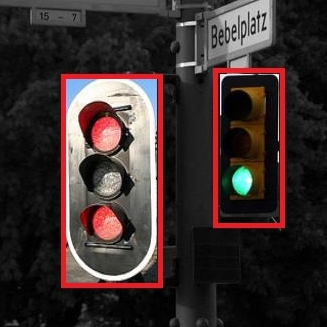}}
\hfill
\subfigure[]{\label{fig:sub-third}\includegraphics[width=.25\textwidth]{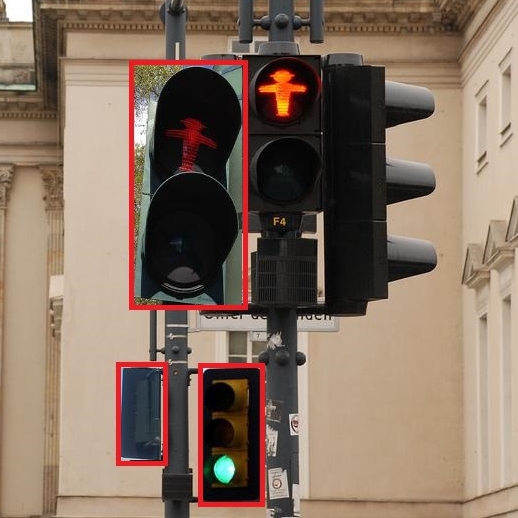}}
\hfill

\subfigure[]{\label{fig:sub-fourth}\includegraphics[width=.25\textwidth]{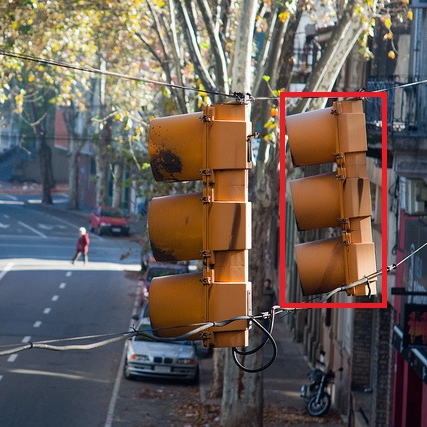}}
\hfill
\subfigure[]{\label{fig:sub-fourth}\includegraphics[width=.25\textwidth]{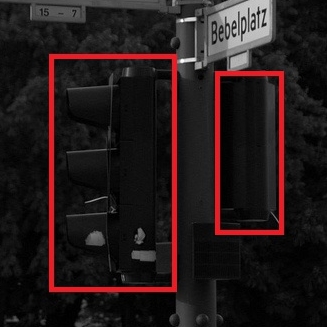}}
\hfill
\subfigure[]{\label{fig:sub-fourth}\includegraphics[width=.25\textwidth]{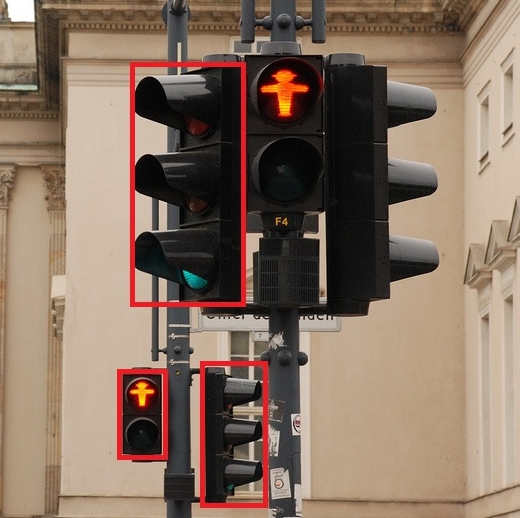}}
\hfill
\caption{\textbf{Row one:} augmented traffic-light images \textbf{Row two:} the original images}
\label{fig:traffic-lights-examples}
\end{figure}
\begin{table}[!htbp]
\label{table:traffic-lights-overall}
\centering
\caption{Baseline Model mAP on \textbf{Traffic Lights}}
\begin{tabular}{l|l|l|l}
                         & C75            & C50            & Loc (C10)      \\ \hline
No Flipping              & 0.096          & 0.363          & 0.539          \\
Flipping                 & 0.099          & 0.366          & 0.553          \\
\textbf{Ours}            & \textbf{0.116}          & \textbf{0.377}          & \textbf{0.534}          \\
\textbf{Flipping + Ours} & \textbf{0.116} & \textbf{0.379} & \textbf{0.547}
\end{tabular}
\end{table}
\begin{table}[!htbp]
\label{table:traffic-lights-large}
\centering
\caption{Baseline Model mAP on \textbf{Traffic Lights (Large Scale)}}
\begin{tabular}{l|l|l|l}
                         & C75            & C50            & Loc (C10)      \\ \hline
No Flipping              & 0.431          & 0.658          & 0.865          \\
Flipping                 & 0.463          & 0.687          & 0.853          \\
\textbf{Ours}            & \textbf{0.460}          & \textbf{0.656}          & \textbf{0.833}          \\
\textbf{Flipping + Ours} & \textbf{0.566} & \textbf{0.727} & \textbf{0.896}
\end{tabular}
\end{table}
\begin{table}[!htbp]
\label{table:traffic-lights-medium}
\centering
\caption{Baseline Model mAP on \textbf{Traffic Lights (Medium Scale)}}
\begin{tabular}{l|l|l|l}
                         & C75            & C50            & Loc (C10)      \\ \hline
No Flipping              & 0.265          & 0.683          & 0.811          \\
Flipping                 & 0.316          & 0.682          & 0.823          \\
\textbf{Ours}            & \textbf{0.258}          & \textbf{0.660}          & \textbf{0.837}          \\
\textbf{Flipping + Ours} & \textbf{0.298} & \textbf{0.686} & \textbf{0.825}
\end{tabular}
\end{table}
\begin{table}[!htbp]
\label{table:traffic-lights-small}
\caption{Baseline Model mAP on \textbf{Traffic Lights (Small Scale)}}
\centering
\begin{tabular}{l|l|l|l}
                         & C75            & C50            & Loc (C10)      \\ \hline
No Flipping              & 0.050          & 0.291          & 0.478          \\
Flipping                 & 0.046          & 0.299          & 0.497          \\
\textbf{Ours}            & \textbf{0.080}          & \textbf{0.316}          & \textbf{0.466}          \\
\textbf{Flipping + Ours} & \textbf{0.065} & \textbf{0.311} & \textbf{0.485}
\end{tabular}
\end{table}


\section{Discussion and Analysis}
As the main metrics of evaluating object detection models, mAP provides an overview of model performance. 
However, for different category instances, the mAP per category varies a lot. 
A normal case is that, categories with large amount of instances comes with higher mAP such as the ``person" category with 44.4\% mAP (C75) containing 262,645 instances in 64,115 images. 
While those with small instance amounts have much lower mAP such as ``hair drier" 1\% mAP (C75) with only 198 instances in 189 images. 
Previous section has presented the experiments of applying slot-based augmentation method into three different category instances. 
These experimental results have shown the benefits of applying augmentation methods. 
In general, applying image augmentation is expected a higher overall detection mAP. 
However, the augmentation affects differently over categories and sometimes it even decreased the detection mAP, such as applying flipping method in boat objects. 
Compared with flipping images, slot-based augmentation produced closing mAP improvement. 
For example, the augmentation on car objects behaved even better than flipping with an extra 2\% mAP. 
Besides, the slot-based augmentation added less images comparing with flipping which doubled the image amount. 
For a large dataset such as MS-COCO, flipping actually increased a large number of images causing extra training time and computation resources. 
While slot-based augmentation method is relatively more flexible to control the amount of augmentation with less increased computation time. 
As the performance for some categories is not improved, combining these methods together produced a consistent improvement over all these three categories. 
For the large boat instance, both flipping and slot-based augmentation failed to improve mAP while the combination provides a 8.6\% increase. 
\newline 

In conclusion, image augmentation is frequently applied in object detection based tasks to provide extra learn-able features and improve the detection mAP. 
One of the commonly used method is flipping all the images of the dataset. 
In this Chapter, a slot-based image augmentation method is proposed, in which images are augmented by replacing isolated foregrounds to provide extra combinations of foreground and backgrounds. 
Additionally, the system components are tested by a series of filter experiments and an augmentation scenario is presented showing detailed procedure of applying this method. 
Besides the expected mAP improvement with augmentation methods, the experimental results have highlighted several interesting facts: 
\begin{itemize}
    \item Instance amounts affects category-wise mAPs. Some categories with lower mAP is resulted from lacking of data, which highlights the importance of applying image augmentation methods on those specific images without rising extra training time. In addition, detailed mAP and the Derek P-R curve is descriptive on analysing detailed performance. 
    \item Image augmentation is not always bringing benefits to detection performance while decreased the mAP instead such as the experimental results of boat instances. 
    \item According to experimental results, the slot-based augmentation approach performed closed behaviour as image flipping with less increased images. Besides slot-based augmentation has large potential to generate images which is a controllable progress by customising slot match ratio and the iteration amount. 
    \item The combination of flipping and slot-based method is relatively more robust and contributes a higher mAP. 
\end{itemize}
Despite those benefits of slot-based augmentation methods, there is a notable effect on contextual features. 
In this chapter, augmentation methods are discussed on improving detection performance while there are many DNN architectural works aiming at extracting the features between foreground and background to enhance detection performance, such as the relation model proposed by \cite{hu2018relation}. 
Slot substitution crops original foreground and replace it with a new one which changes the foreground-background relational features. 
However, this type of changes has critical effects. 
On one hand, breaking the original contextual features makes it difficult for DNN models to detect objects according to their surroundings. 
On the other hand, breaching original contextual features potentially contributes a robust feature extraction that recognising objects fully by its features not by the environments. 
For example, human beings recognise the flying beer on the sky from advertisement posters. 


\bibliographystyle{unsrt}  
\bibliography{references}

\end{document}